\documentclass[letterpaper, 10 pt, conference]{ieeeconf}  
\usepackage{iftex,amsmath}
\iftutex
\usepackage{unicode-math}
\else
\usepackage{amssymb}
\usepackage{graphicx,subfigure}
\newcommand{\norm}[1]{\left\lVert#1\right\rVert}
\usepackage{upgreek}
\usepackage{stackengine}
\def\delequal{\mathrel{\ensurestackMath{\stackon[1pt]{=}{\scriptstyle\Delta}}}}
\usepackage{multicol}
\usepackage{tablefootnote}

\usepackage{amsthm}

\IEEEoverridecommandlockouts                              

\title{\LARGE \bf
Robust Artificial Delay based Impedance Control of Robotic Manipulators with Uncertain Dynamics \\
}

\author{Udayan Banerjee$^{1}$, Bhabani Shankar Dey$^{2}$, Indra Narayan Kar$^{3}$, Subir Kumar Saha$^{4}$
\thanks{*This work was not supported by any organization}
\thanks{$^{1}$Udayan Banerjee is a PhD student in the School of     Interdisciplinary Research, Indian Institute of Technology Delhi, Hauz Khas, New Delhi, 110016, India
        {\tt\small srz198482@iitd.ac.in}}%
\thanks{$^{2}$Bhabani Shankar Dey is a PhD student in the Department of Electrical Engineering, Indian Institute of Technology Delhi, Hauz Khas, New Delhi, 110016, India
        {\tt\small bhabanishankar440@gmail.com}}%
\thanks{$^{4}$Subir Kumar Saha is a faculty in the Department of Mechanical Engineering, Indian Institute of Technology Delhi, Hauz Khas, New Delhi, 110016, India
        {\tt\small saha@mech.iitd.ac.in}}%
\thanks{$^{3}$Indra Narayan Kar is a faculty in the Department of Electrical Engineering, Indian Institute of Technology Delhi, Hauz Khas, New Delhi, 110016, India
        {\tt\small ink@ee.iitd.ac.in}}}%

\usepackage{amsmath,amsfonts,amsthm,bm} 

\begin{document}

\maketitle
\thispagestyle{empty}
\pagestyle{empty}

\begin{abstract}
In this paper an artificial delay based impedance controller is proposed for robotic manipulators with uncertainty in dynamics. The control law unites the time delayed estimation (TDE) framework with a second order switching controller of super twisting algorithm (STA) type via a novel generalized filtered tracking error (GFTE). While time delayed estimation framework eliminates the need for accurate modelling of robot dynamics by estimating the uncertain robot dynamics and interaction forces from immediate past data of state and control effort, the second order switching control law in the outer loop provides robustness against the time delayed estimation (TDE) error that arises due to approximation of the manipulator dynamics. Thus, the proposed control law tries to establish a desired impedance model between the robot end effector variables i.e. force and motion in presence of uncertainties, both when it is encountering smooth contact forces and during free motion. Simulation results for a two link manipulator using the proposed controller along with convergence analysis are shown to validate the proposition.
\end{abstract}
\section{INTRODUCTION}
\subsection{Background and Motivation}
With the advent of safety focused collaborative robots in real world applications, research in the area of interaction control and physical-human-robot-interaction (pHRI) has received significant attention. In order to achieve a stable interaction between robot and the human in a shared work-space, direct or indirect regulation of the contact forces becomes essential. Hence, researchers in the past had focused on either controlling position \cite{gasparetto2012automatic} or end point force of robotic manipulator. The hybrid position and force control was later introduced in \cite{raibert1981hybrid}, \cite{mason1981compliance} where a combination of end effector position and force variables are controlled in orthogonal subspaces, while \cite{stepien1987control} showed the effectiveness of force control for interactive tasks performed in a structured environment. These traditional algorithms work well for robotic applications where the robot end effector is always in contact with the environment but \cite{whitney1977force} \cite{colgate1989analysis} showed that even in situations where the contact is guaranteed, these force control algorithms can cause stability issues.\par
Later on \cite{hogan1984impedance}\cite{hogan1985impedance} proposed that instead of controlling the end point position and force variables independently, modulating the dynamic relationship between these variables at the interaction port can ensure a stable dynamic interaction. This method is popularly termed as "Impedance Control" in literature where the end point impedance of the robot is tuned to obtain a stable response to a reaction force. However, for practical implementations impedance controllers are strongly affected by uncertainties associated with the coupled non-linear robot dynamics, unknown external environment and disturbances. Several control strategies have been reported in literature that address the robustness issue of impedance controllers. In \cite{chan1991robust} variable structure based robust impedance controllers have been proposed whereas \cite{ibeas2004robust} used sliding mode controllers to tackle modelling errors and uncertainties while implementing the impedance model. Various adaptive impedance controllers \cite{kelly1989adaptive}\cite{lu1991impedance} have also been proposed in this regard where both modelling uncertainties in robot dynamics and adaptability to unknown environment \cite{zhang2016development}\cite{colbaugh1993direct} have been addressed through adaptation of impedance parameters. \par Most of these impedance controllers require precise modelling of the robot dynamics and are of complex structure  demanding high computational power. In this front, the framework of artificial delay based or time delayed control (TDC) has shown great potential for motion control systems. It offers a model free control structure where a synthetic delay is injected in the closed loop system to estimate the robot dynamics using the data of control input and state of previous sampling instant \cite{roy2020adaptive}. It is not only easy to implement, but demands less computational power by eliminating the need of complex modelling. Artificial delay based impedance control or time delayed impedance control (TDIC) has been used in \cite{souzanchi2017robust} and\cite{jung2021similarity} but the negotiation of time delayed estimation error (TDE) which can cause serious degradation of the close loop performance \cite{roy2019new} and stability issues \cite{roy2017adaptive} has not been addressed in any of these studies.
\par                                          
In most of the existing literature \cite{roy2020adaptive}, time delayed estimation error is mitigated by robustifying  outer loop using first order switching while \cite{kali2018super} and \cite{tran2021novel} have used super twisting algorithms to tackle the estimation error in motion control problems. Reasoning for using switching term to deal with perturbed cases has been an established result in the sliding mode control literature \cite{shtessel2014sliding}. But most of these studies adopt conservative assumptions on the TDE error which limits the applicability of this framework to  a larger class of systems. 
\subsection{Contributions}
In this regard, this paper proposes a new robust artificial delay based impedance controller with following new salient points
\begin{itemize}
     \item Introduction of a generalized filtered tracking error for constrained robot motion tasks.
    \item Integral switching surface design using an LTI representation of the impedance model error.
    \item Introduction of a STA type second order switching control in the outer loop of the controller to tackle TDE error. 
    \item Proposed design uses a state dependent upper-bound structure of the uncertainty unlike in \cite{kali2018super} and \cite{tran2021novel} where constant upper-bounds were used.
    \item Compensation of TDE error during free and constrained motion of the robot with the same auxilliary control.
\end{itemize}
In section (\ref{Dynamic Model}), the dynamics of $n$ degree of freedom rigid manipulator is elucidated. Design of robust artificial delay based impedance controller is introduced in section (\ref{Robust artificial Delay based Impedance Controller}) followed by the stability analysis of the closed loop system in section (\ref{Stability Analysis}). To corroborate the propositions, section (\ref{Simulation Studies}) illustrates the simulation results followed by the concluding remarks in section (\ref{Conclusions}).
\section{Dynamic Model}\label{Dynamic Model}
Consider the Cartesian space dynamics of an $n$ degree of freedom constrained rigid robot manipulator expressed in (\ref{eq1}), where ${x,\dot{x},\ddot{x}\in \mathbb{R}^n}$ denote the position, velocity and acceleration vectors in Cartesian space respectively, ${M_x\left(q\right)}\in \mathbb{R}^{n\times n}$ signifies the Cartesian inertia matrix, ${C_x(q,\dot{q})}\in \mathbb{R}^{n\times n}$ denotes the Cartesian Centripetal-Coriolis matrix, $ g_x\left(q\right)\in \mathbb{R}^n$ denotes the Cartesian gravity vector, $d_x\left(t\right)\in \mathbb {R}^n$ captures the lumped disturbances, $F_e\in \mathbb{R}^n$ encapsulates the measurable interaction forces acting on the end effector and $F_u\in \mathbb{R}^n$ represents the control effort in Cartesian space.
\begin{equation}
\label{eq1}
{{M_x\left(q\right)\ddot{x}+C_x\left(q,\dot{q}\right)\dot{x}+g_x\left(q\right)+d_x}}(t)={F_u(t)+F_e(x)}
\end{equation}
\begin{equation}
\label{eq2}
{{M}_x(q){\ddot{x}+H(x,\dot{x})}}={F_u}(t)
\end{equation}
Manipulator dynamics in (\ref{eq1}) can be remodified to (\ref{eq2}) with ${H(x,\dot{x}) \delequal C_x\left(q,\dot{q}\right)\dot{x}+g_x\left(q\right)+d_x}\left(t\right)-{F_e(x)}$ where $q,\dot{q},\ddot{q}\in \mathbb{R}^n$ denote joint position, velocity and acceleration vectors respectively. Joint variables can be related to their Cartesian counterparts using (\ref{eq3}) and (\ref{eq4}) where $J_a(q)\in \mathbb{R}^{n\times n}$ signifies the analytical Jacobian of the manipulator.
\begin{equation}
\label{eq3}
{\dot{x}=J_a\dot{q}}
\end{equation}
\begin{equation}
\label{eq4}
{\ddot{x}=J_a\ddot{q}+{\dot{J}}_a\dot{q}}
\end{equation}
Inertia matrix, Coriolis matrix and gravity vector in (\ref{eq1}) can be obtained using (\ref{eq5})-(\ref{eq7}), where $M(q),C(q,\dot{q})\in \mathbb{R}^{n\times n}$ and $g(q)\in \mathbb{R}^{n}$ denote the joint space components.
\begin{equation}
\label{eq5}
  {M_x(q) = J_a^{-T}(q)M(q)J_a^{-1}(q)}
\end{equation}
\begin{equation}
\label{eq6}
\begin{split}
    {C_x(q,\dot{q}) = J_a^{-T}(q)C(q,\dot{q})-M_x(q)\dot{J}_a(q)J_a^{-1}(q)}
     \end{split}
\end{equation}
\begin{equation}
\label{eq7}
   {g_x(q) = J_a^{-T}g(q)}
\end{equation}
Further, control torque in joint coordinates is given by (\ref{eq8}).
\begin{equation}
\label{eq8}
{\tau}={J_a^TF_u}
\end{equation}
\textit{Property 1}: Cartesian inertia matrix $M_x(q)$ and its inverse $M_x^{-1}(q)$ are uniformly positive definite and $\exists$ $\beta_1,\beta_2\in \mathbb{R^+}$ such that the following inequalities hold.
\begin{equation}
\label{eq9}
   \beta_1I\leq M_x(q)\leq\beta_2I
\end{equation}
\textit{Assumption 1}: Manipulator is working in a singularity free region of the task space.
\section{Robust artificial Delay based Impedance Controller}\label{Robust artificial Delay based Impedance Controller}
\subsection{Cartesian Impedance Control}
The goal of classical impedance controller is to realize a desired dynamic behavior between robots force and motion variables in Cartesian space as shown below,
\begin{equation}
\label{eq10}
    {M_m(\ddot{x}-\ddot{x}_d)+D_m(\dot{x}-\dot{x}_d)+K_m(x-x_d) = F_e}
\end{equation}
where $M_m,D_m,K_m\in\mathbb{R}^{n\times n}$ denote the desired inertia, damping and stiffness matrices with diagonal entries and $x_d$,$\dot{x}_d,\ddot{x}_d \in \mathbb{R}^{n}$ representing the desired position, velocity and acceleration trajectories respectively, while $F_e$ signifies the interaction force \cite{chan1991robust}. The desired trajectory $x_d$ is chosen slightly inside the environment to ensure contact. 
\subsection{Impedance Error Representation using a GFTE}
Defining the position error as ${e = x_d-x}$ and using (\ref{eq10}) the Cartesian impedance error $\eta$ can be constructed as (\ref{eq11}) where the gains $K_p$, $K_d$ are related to the impedance parameters as ${K_p=M_m^{-1}K_m}$ and ${K_d=M_m^{-1}D_m}$.  
\begin{equation}
\label{eq11}
{\eta}\delequal{\ddot{e}+K_d\dot{e}+K_pe+M_m^{-1}F_e}
\end{equation}
Let ${\alpha}\in\mathbb{R}^n$ in (\ref{eq12}) denote a Generalized Filtered Tracking Error (GFTE) that is considered as a sum of velocity error $\dot{{e}}\in\mathbb{R}^n$, position error ${e}\in\mathbb{R}^n$ and an auxilliary error variable ${e_f}\in\mathbb{R}^n$, such that the interaction force $F_e$ and the auxiliary error ${e_f}$ obey a first order relation as given in (\ref{eq13}) where ${\Gamma_1,\Gamma_2}\in\mathbb{R}^{n\times n}$ denote positive definite diagonal matrices.
\textit{\begin{equation}
\label{eq12}
{\alpha} \delequal{\dot{{e}}+{\Gamma_1}{e}+{e_f}}
\end{equation}
\begin{equation}
\label{eq13}
{\dot{{e}}_f}+{\Gamma_2}{e_f}={M_m^{-1}F_e}
\end{equation}}
Impedance error ${\eta}$ in (\ref{eq11}) can now be expressed in terms of the GFTE following a linear time invariant (LTI) representation as shown in (\ref{eq14}) where 
${K_p} = {\Gamma_2\Gamma_1}$ and ${K_d} = {\Gamma_1+\Gamma_2}$.
\begin{equation}
\label{eq14}
\eta=\dot{\alpha}+\Gamma_2\alpha
\end{equation}
Thus, a switching surface ${s}$ is selected as (\ref{eq15}) where $s \in \mathbb{R}^n$.
\begin{equation}
\label{eq15}
{s} = \int_{0}^{t}{\eta~dt}
\end{equation}
\theoremstyle{remark}
\newtheorem{remark}{Remark}
\begin{remark}
 It can be appreciated that in the absence of any interaction force $F_e$, the auxilliary variable $e_f$ vanishes from $\alpha$ and the GFTE simplifies to the conventional filtered tracking error used for a position tracking problem.
\end{remark}
\textit{Assumption 2}: The contact force model is assumed to be of viscoelastic type, hence $F_e \delequal K_e(x_e-x)$ where $x_e\in\mathbb{R}^{n}$ indicates the location of the compliant environment and $K_e\in\mathbb{R}^{n\times n}$ denotes the environmental stiffness.

\subsection{Robust Artificial Delay based Impedance Control law}
Modified robot dynamics in (\ref{eq2}) can be reformulated in a more compact way using a positive definite design matrix ${{\bar{M}}_x}\in \mathbb{R}^{n\times n}$ with constant elements as shown in (\ref{eq16}) where $N$ as indicated in (\ref{eq17}) captures uncertain robot dynamics \cite{roy2017adaptive}.
\begin{equation}
\label{eq16}
{\bar{M}}_x\ddot{x}+N(x,\dot{x},\ddot{x})=F_u
\end{equation}
\begin{equation}
\label{eq17}
N\left(x,\dot{x},\ddot{x}\right)=\left[M_x\left(q\right)-{\bar{M}}_x\right]\ddot{x}+H
\end{equation}
The function ${N}$ is approximated with $\hat{N}$ and it is calculated using the past measurements of state and input as shown in (\ref{eq18}) where $h$ is an artificially introduced small delay.
\begin{equation}
\label{eq18}
{N\cong \hat{N} = {F_u}\left(t-h\right)-{{\bar{M}}_x\ddot{x}}(t-h)}
\end{equation}
Based on the surface defined in (\ref{eq15}), the robust artificial delay based impedance control law is proposed as (\ref{eq19}) where the auxilliary control $a$ is selected as (\ref{eq20}) and the robustifying part $\Delta a$ is chosen as the second order switching controller defined in (\ref{eq21}),
\begin{equation}\label{eq19}
{F_u={\bar{M}}_x}({a}+{\Delta} {a})+{F_u}\left(t-h\right)-{{\bar{M}}_x\ddot{x}}\left(t-h\right)
\end{equation}
\begin{equation}
\label{eq20}
    {a = \ddot{x}_d+\Gamma_1\dot{e}+\dot{e}_f+\Gamma_2\alpha}
\end{equation}
\begin{equation}
\begin{array}{ll}
\label{eq21}
{\Delta}{a} = {\lambda_{1}}({t})\dfrac{s}{\norm{s}^\frac{1}{2}}
-y\\ \\
{\dot{y}} = -{\lambda_{2}}(t)\dfrac{s}{\norm{s}}
\end{array}
\end{equation}\\
with $\lambda_1,\lambda_2\in\mathbb{R^+}$ denoting time varying gains. Structure of each gain is given by (\ref{eq22}) with $\gamma_{01}^*,\gamma_{02}^*,\gamma_{11}^*,\gamma_{12}^*\in \mathbb{R^+}$ and 
${\Theta}^T = \begin{bmatrix}e^T&\dot{e}^T\end{bmatrix}$.
\begin{equation}
    \begin{array}{ll}\label{eq22}
{\lambda_1(t)} = 2(\gamma_{01}^*+\gamma_{11}^*\norm{{\Theta}})\\ \\ {\lambda_2(t)} = 2(\gamma_{02}^*+\gamma_{12}^*\norm{{\Theta}})
\end{array}
\end{equation}
Finally the joint space impedance control law is given as (\ref{eq23}).
\begin{equation}
\label{eq23}
{\tau}={J_a}^T\left({\bar{M}}_x({a}+{\Delta} {a)}+{F_u}\left(t-h\right)-{\bar{M}}_x\ddot{x}\left(t-h\right)\right)
\end{equation}
\section{Stability Analysis}\label{Stability Analysis}
\subsection{Closed Loop Dynamics}
Applying control law (\ref{eq19}) in the open loop robot dynamics shown in (\ref{eq16}), the closed loop system is obtained as (\ref{eq24}), where ${\sigma}\delequal{{\bar{M}}_x^{-1}}\left({\Delta} {N}\right)$ is the time delayed estimation (TDE) error with ${\Delta}{N}={N-\hat{N}}$, which arises due to approximation of the uncertain robot dynamics from the past measurement of input and state.
\begin{equation}
\label{eq24}
{\ddot{x}} = {a+\Delta a} - {\sigma}
\end{equation}
 Using the relations in (\ref{eq12}) and (\ref{eq13}) the time derivative of the switching surface simplifies to (\ref{eq25}).
\begin{equation}
\label{eq25}
\dot{s}=\ddot{{e}}+{\Gamma_{1}}\dot{e}+\dot{{e}_f}+{\Gamma_{2}}{\alpha}
\end{equation}
Simplifying (\ref{eq25}) using the closed loop dynamics in (\ref{eq24}),
\begin{equation}
\label{eq26}
\dot{s}={\ddot x}_d-({a+\Delta a}-{\sigma})+{\Gamma_1}\dot{e}+{\dot{e}_f}+{\Gamma_2}{\alpha}
\end{equation}
Replacing the auxiliary control input defined in (\ref{eq20}), the surface dynamics reduces to the following form.
\begin{equation}\label{eq27}
    \dot{s}=-{\Delta}{a}+{\sigma}
\end{equation}
Applying the robustifying part as dictated in (\ref{eq21}), the time derivative of the surface modifies as follows,
\begin{equation}\label{eq28}
\begin{array}{ll}
     \dot{s}=-{\lambda_{1}}(t)\dfrac{s}{\norm{s}^\frac{1}{2}}
+y + {\sigma}\\ \\
     {\dot{y}} = -{\lambda_{2}}(t)\dfrac{s}{\norm{s}}
\end{array}
\end{equation}
Using the transformation ${\Psi}={\sigma}+{y}$, where $\Psi \in \mathbb{R}^n$, (\ref{eq28}) can be modified to (\ref{eq29}).
\begin{equation}
    \begin{array}{ll}\label{eq29}
     \dot{s}=-{\lambda_{1}}(t)\dfrac{s}{\norm{s}^\frac{1}{2}}+{\Psi}\\ \\
     {\dot{\Psi} = -{\lambda_{2}}(t)\dfrac{s}{\norm{s}}}+\dot{\sigma}
\end{array}
\end{equation}
\subsection{Upperbound of TDE Error Derivative}
It has been already established in \cite{roy2019new} that if the user defined positive definite matrix ${\bar{M}}_x$ is selected such that the inequality $\Omega\delequal{\norm{{M_x^{-1}\left(q\right){\bar{M}}_x-I}}}<1$ is satisfied at all instants of time, then the TDE error $\sigma$ involved in a TDC framework has a state dependent upperbound structure as indicated in (\ref{eq30}) , where $\gamma_0,\gamma_1\in \mathbb{R^+}$ and ${\Theta}^T = \begin{bmatrix}e^T&\dot{e}^T\end{bmatrix}$.
\begin{equation}
\label{eq30}
   \norm{\sigma}\leq({\gamma_0+\gamma_1}\norm{{\Theta}})
\end{equation}
Derivative of the TDE error can be expressed as (\ref{eq31}) where ${\sigma_h}\delequal{ {\sigma(t-h)}}$. 
\begin{equation}
\label{eq31}
    \dot{\sigma} = \lim_{h\to0}\frac{{{\sigma}} -{{\sigma_h}}}{h}
\end{equation}
Applying triangle inequality on (\ref{eq31}), (\ref{eq32}) is obtained.
\begin{equation}
    \label{eq32}
     \norm{\dot{{\sigma}}} \leq \lim_{h\to0}\frac{\norm{{\sigma}}+\norm{{\sigma_h}}}{h}
\end{equation}
Now considering a sufficiently small delay and using (\ref{eq30}), upperbound on TDE error derivative is obtained as (\ref{eq33}),
\begin{equation}
    \label{eq33}
     \norm{\dot{{\sigma}}} \leq {2h^{-1}(\gamma_0+\gamma_1\norm{{\Theta}}) = 2h^{-1}(\Delta)}
\end{equation}
where $\Delta \delequal (\gamma_0+\gamma_1\norm{{\Theta}})$. Exact information on $\gamma_0$ and $\gamma_1$ will require the instantaneous values of $\Omega$ which is generally not known. But, as ${\Omega}<1$ the upperbounds $\gamma_0\leq\mu/(1-\bar{\Omega})$ and $\gamma_1\leq\bar{\Omega}\norm{K}/(1-\bar{\Omega})$ for scalars $\bar{\Omega}<1$ and $\mu>0$ will always exist where $K\delequal{\begin{bmatrix}{K_p}&{K_d}\end{bmatrix}}$\cite{roy2019new}. This will be further used for stability analysis.
\subsection{Lyapunov Analysis}
\textit{Theorem 1}: Let $V(.)$ be a continuously differentiable candidate Lyapunov function with state ${\Phi}^T= \begin{bmatrix}\phi^T_1(s)& {\Psi}^T\end{bmatrix}$ and $\phi_1(s) = \dfrac{s}{\norm{s}^{\frac{1}{2}}}$ such that,
\begin{equation}
\label{eq34}
  {V}(\Phi) = \dfrac{1}{2}{\Phi}^TP\Phi
  \end{equation}
and,
\begin{equation}\label{eq35}
    \dot V\leq-\rho(t) \sqrt{V},  for \norm{\Theta}>\epsilon\Rightarrow \norm{s}>\epsilon^*
\end{equation}
 For a matrix ${Q(t)>0}$, if $\exists$ a matrix $P(t)>0$ satisfying the matrix differential Riccatti (MDRE) type equation in (\ref{eq36}) and for the Lyapunov function in (\ref{eq34}), condition stated in (\ref{eq35}) holds true, then solutions of system (\ref{eq29}) uniformly converge to a bound $\epsilon^* \in \mathbb{R^+}$.
\begin{equation}\label{eq36}
    \norm{{s}}^{\frac{1}{2}}(\dot{P}+PBB^TP)+A^TP+PA+ \norm{{s}}^{-\frac{1}{2}}Q=0
\end{equation}\\
\textit{Proof}:
Using (\ref{eq29}), the state dynamics corresponding to vector $\Phi$ is given by (\ref{eq37}) where $A=(\bar{A}\otimes \phi_2(s)) \in\mathbb{R}^{2n\times 2n}$, $B=(\bar{B}\otimes I_n) \in\mathbb{R}^{2n}$ and $\phi_2(s) =  \norm{{s}}^{\frac{1}{2}}\dfrac{\delta\phi_1(s)}{\delta s} \in\mathbb{R}^{n\times n}$,
\begin{equation}
\label{eq37}
     \dot{{\Phi}} = \norm{{s}}^{-\frac{1}{2}}(A\Phi+B\norm{s}^{\frac{1}{2}}\dot{\sigma})
 \end{equation}
with $\bar{A} = \begin{bmatrix}
-{\lambda_1(t)} & {1}\\[.25cm]
-2{\lambda_2(t)} & {0}
\end{bmatrix}$, $\bar{B} = \begin{bmatrix}{0} \vspace{0.2cm}\\{1}\end{bmatrix}$.\\\\
Time derivative of the Lyapunov function (\ref{eq34}) along state trajectories yields,
\begin{equation}
\label{eq38}
    \begin{split}
    {\dot{V}} = \frac{1}{2}\norm{{s}}^{-\frac{1}{2}}\{{\Phi}^T(A^TP+PA+\norm{{s}}^{\frac{1}{2}}\dot{P})\Phi\\+ 2\norm{s}^{\frac{1}{2}}\dot{\sigma}^TB^TP\Phi\}
    \end{split}
\end{equation}
For any scalar $\delta>0$ and a matrix ${D = I>0}$, following inequality holds,
\begin{equation}\label{eq39}
   2\norm{s}^{\frac{1}{2}}\dot{\sigma}^TB^TP\Phi\leq(\delta\dot{\sigma}^T\dot{\sigma}+\frac{1}{\delta}\Phi^TPBB^TP\Phi)\norm{s}^{\frac{1}{2}}
\end{equation}
Using above inequality, (\ref{eq38}) can be modified as, 
\begin{equation}\label{eq40}
     \begin{split}
      \dot{V}\leq\frac{1}{2}\norm{{s}}^{-\frac{1}{2}}\{{\Phi}^T(A^TP+PA+\norm{{s}}^{\frac{1}{2}}\dot{P}){\Phi}+\\\delta\dot{{\sigma}}^T\dot{{\sigma}}\norm{{s}}^{-\frac{1}{2}}\norm{{s}}
    +\frac{1}{\delta}{\Phi}^TPBB^TP\Phi\norm{{s}}^{\frac{1}{2}}\}
    \end{split}
\end{equation}
Simplifying (\ref{eq40}) with matrix $M \delequal C^TC$ and $C\delequal \begin{bmatrix} {I_n} & 0\end{bmatrix}$,
\begin{equation}\label{eq41}
\begin{split}
      \dot{V}\leq\frac{1}{2}\norm{s}^{-\frac{1}{2}}\{\Phi^T(A^TP+PA+\norm{{s}}^{\frac{1}{2}}\dot{P})\Phi+\\\delta\norm{\dot{\sigma}}^2\Phi^TM\Phi\norm{{s}}^{-\frac{1}{2}}
    +\frac{1}{\delta}\norm{{s}}^{\frac{1}{2}}\Phi^TPBB^TP\Phi\}
    \end{split}
\end{equation}
Further, applying the upperbound of ${\dot{\sigma}}$ from (\ref{eq33}), 
\begin{equation}\label{eq42}
\begin{split}
      \dot{V}\leq \frac{1}{2}\norm{s}^{-\frac{1}{2}}{\Phi}^T\{\norm{{s}}^{\frac{1}{2}}(\dot{P}+\frac{1}{\delta}PBB^TP)+A^TP+PA+\\\norm{{s}}^{-\frac{1}{2}}\frac{4\delta\Delta^2}{h^2}M\}{\Phi}
    \end{split}
\end{equation}
With $\delta=1$ and using (\ref{eq36}), $\dot{V}$ can be upper-bounded as,
\begin{equation}\label{eq43}
\begin{split}
      \dot{V}\leq -\frac{1}{2}\norm{s}^{-\frac{1}{2}}\Phi^T\left(Q-\frac{4\Delta^2}{h^2}M\right)\Phi
    \end{split}
\end{equation}
\\
Let, $\tilde{Q} \delequal {Q}-\dfrac{4\Delta^2}{h^2}{M}$, \\
\begin{equation}\label{eq44}
      \dot{V}\leq -\frac{1}{2}\norm{s}^{-\frac{1}{2}}{\Phi}^T\{\tilde{Q}\}{\Phi} 
\end{equation}
For the ease of analysis, a region of the state space defined by the variables $e, \dot{e}$ is considered as shown in (\ref{eq45}), where $\epsilon \delequal {\dfrac{1}{\gamma_0^*-\gamma_1^*}}$, with $\gamma_0^*,\gamma_1^*\in \mathbb{R^+}$ and $\gamma_0^*>\gamma_1^*$.
\begin{equation}\label{eq45}
    \norm{\Theta}>\epsilon \Rightarrow \norm{\Theta}>\frac{1}{(\gamma_{0}^*-\gamma_{1}^*)}
\end{equation}
Matrix $Q$ is designed using ${\lambda_1,\lambda_2}>0$, artificial delay $h$ and a positive constant $\epsilon$ as shown in (\ref{eq46}). It is positive definite in the region (\ref{eq45}) for any $P(t)>0$ satisfying (\ref{eq36}).
\begin{equation}\label{eq46}
    {Q} = \begin{bmatrix}
\dfrac{({\lambda_1}+{\lambda_2)}^2}{h^2}+\dfrac{{\lambda_1}^2\norm{\Theta}^2}{h^2}& \dfrac{-{\lambda_1}\epsilon}{{h}}\\[.35cm]
\dfrac{-{\lambda_1}\epsilon}{h} & {1}
\end{bmatrix} \otimes I_n
\end{equation}
Further, ${\tilde{Q}}$ can be obtained from $Q$ as shown in (\ref{eq47}).

\begin{equation}\label{eq47}
    \tilde{{Q}} = \begin{bmatrix}
\dfrac{({\lambda_1}+{\lambda_2)}^2}{h^2}+\dfrac{{\lambda_1}^2\norm{\Theta}^2}{h^2}-\dfrac{4{\Delta}^2}{h^2}& \dfrac{-{\lambda_1}\epsilon}{{h}}\\[.35cm]
\dfrac{-{\lambda_1}\epsilon}{h} & {1}
\end{bmatrix} \otimes I_n
\end{equation}
and it is also positive definite in the region (\ref{eq45}) if ${\lambda_1,\lambda_2}$ are selected as in (\ref{eq22}) and the coefficients are chosen such that $\gamma_0^*>\gamma_1^*$, $\gamma_{0}^*=(\gamma_{01}^*+\gamma_{02}^*)>\gamma_0$, $\gamma_{1}^*=(\gamma_{11}^*+\gamma_{12}^*)> \gamma_1$. 
\\ \\
$\dot{V}$ can be further simplified to (\ref{eq48}) from (\ref{eq44}) as,
\begin{equation}\label{eq48}
      \dot{V} \leq -\frac{1}{2} \norm{{s}}^{-\frac{1}{2}} \lambda_{min}(\tilde{Q})\norm{{\Phi}}^2 < 0
\end{equation}
\begin{figure*}[h!]
\centering
      \subfigure[Impedance error in x-axis \label{fig1:a}
      $\left(\dfrac{m}{s^2}\right)$]{\includegraphics[width=.329\textwidth]{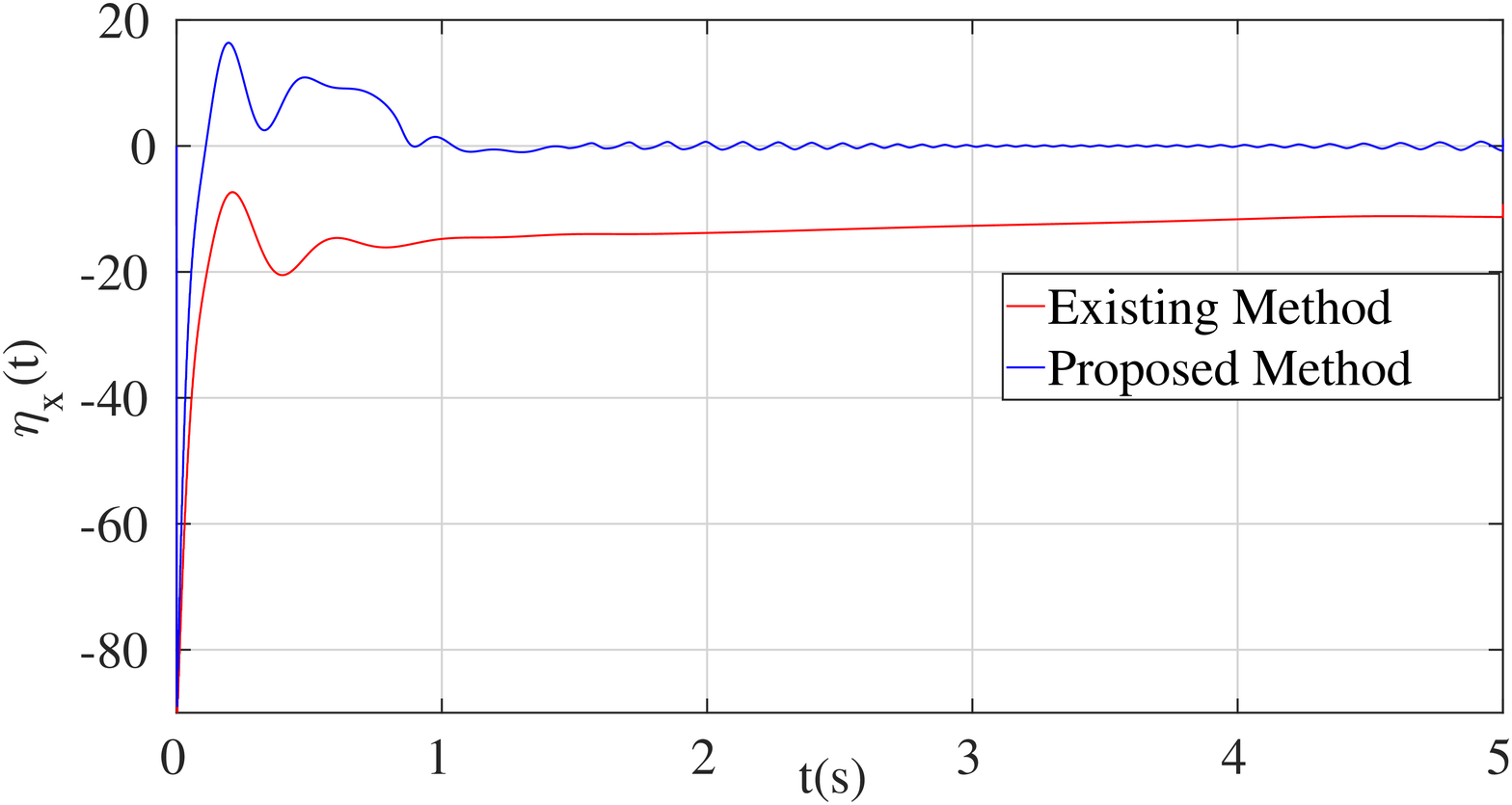}}
      \subfigure[Impedance error in y-axis \label{fig1:b}
      $\left(\dfrac{m}{s^2}\right)$]{\includegraphics[width=.329\textwidth]{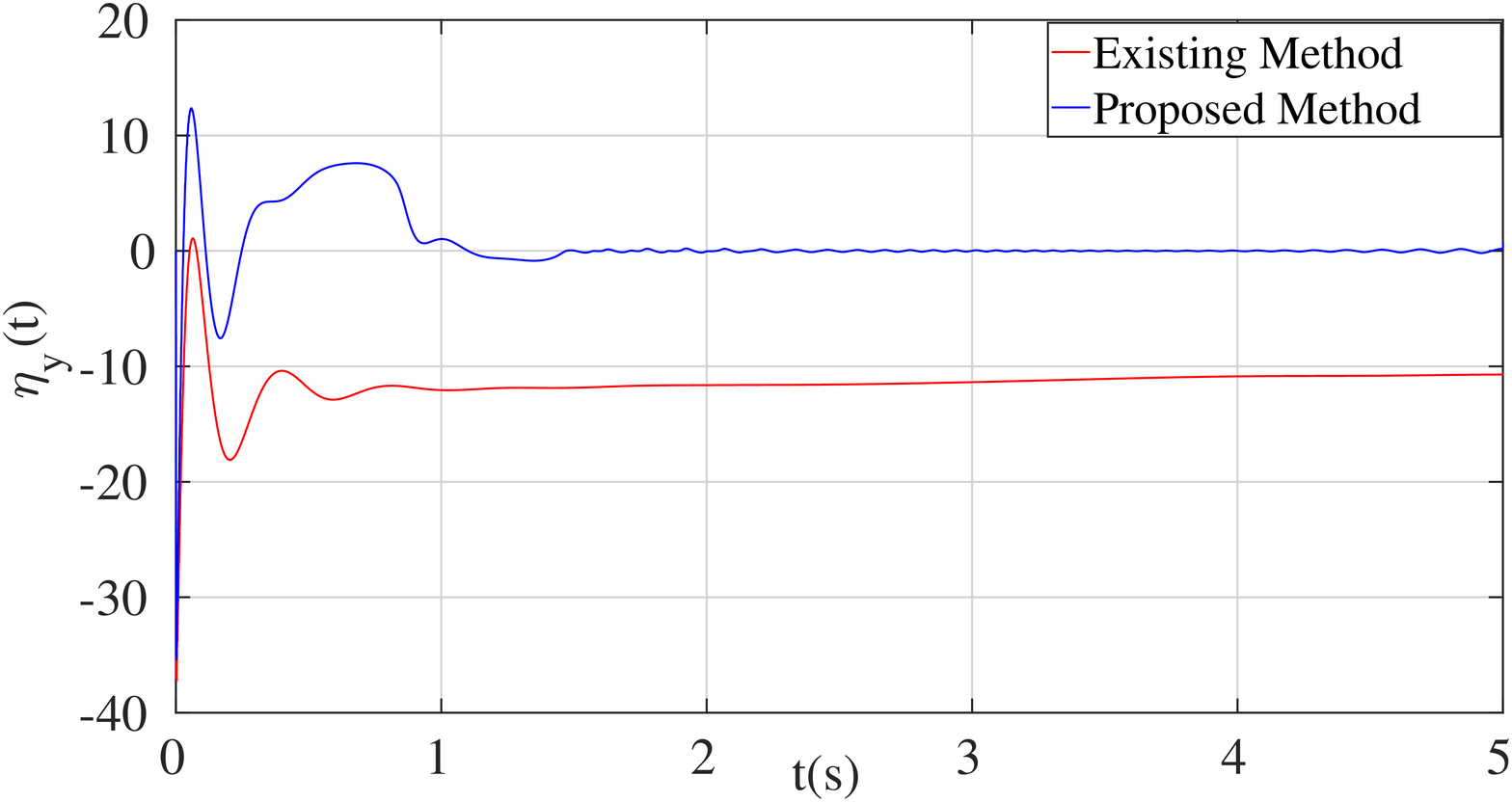}}
          \subfigure[Position error in x-axis \label{fig1:c} ($m$)]{\includegraphics[width=.329\textwidth]{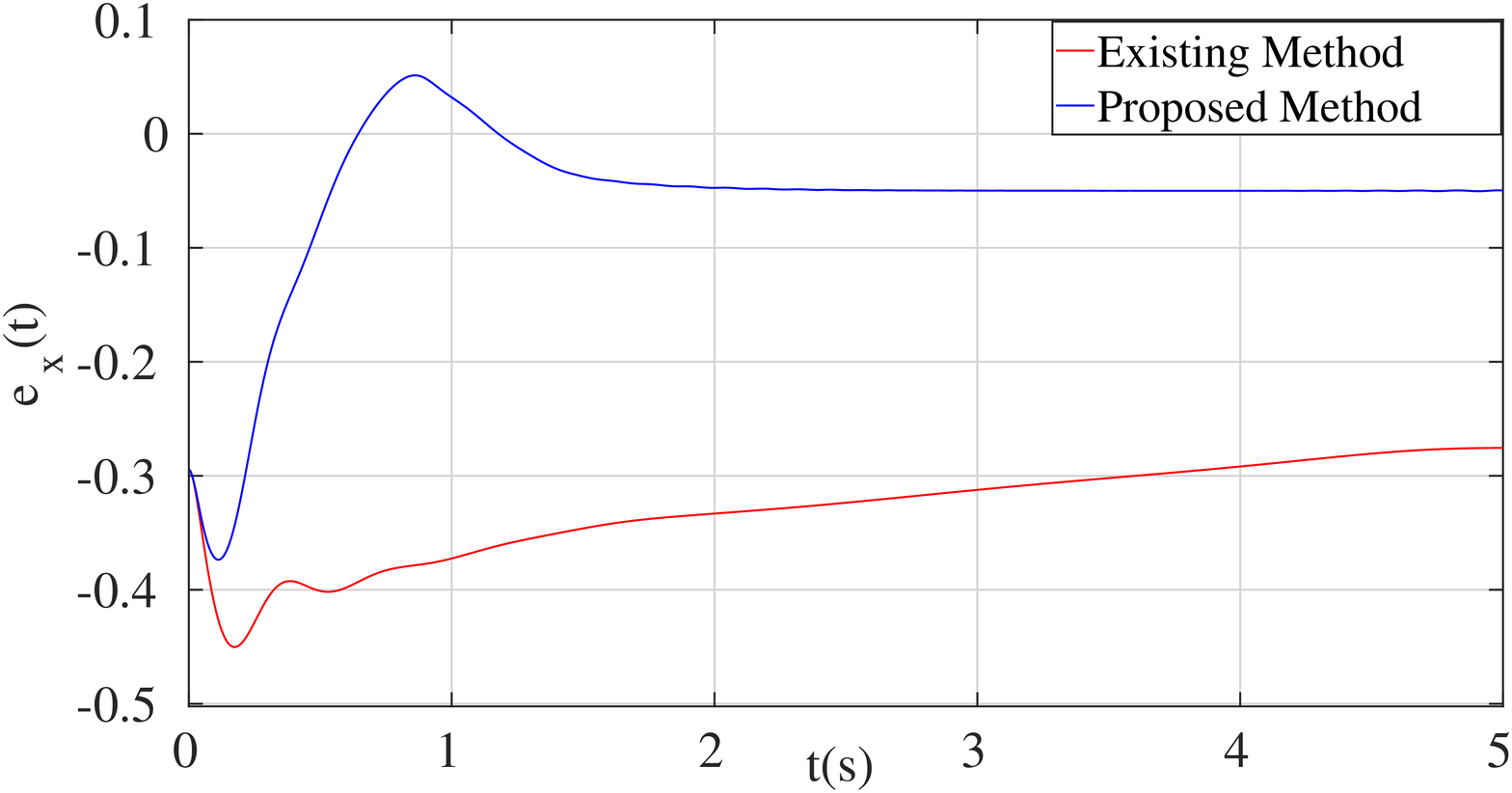}}
       \caption{Comparative Results of Proposed Robust Artificial Delay based Impedance Controller and Existing TDIC}
       \label{fig1}
      \end{figure*}
 which indicates that it is negative definite in the region (\ref{eq45}) and solutions of system in (\ref{eq29}) uniformly converge to the bound $\epsilon^*$. Applying Rayleigh's inequality on (\ref{eq34}) and taking norm of the vector ${\Phi}$ defined in the beginning of Lyapunov analysis, following inequalities are obtained.
\begin{equation}\label{eq49}
     \lambda_{min}(P) \norm{\Phi}^2 \leq V(\Phi) \leq  \lambda_{max}(P) \norm{\Phi}^2
\end{equation}
\begin{equation}\label{eq50}
    \norm{\Phi}^2=\norm{s}+ \norm{\Psi}^2 \Rightarrow \norm{s}^{\frac{1}{2}} \leq \norm{\Phi} \leq \left[\frac{V(\Phi)}{\lambda_{min}(P)}\right]^{\frac{1}{2}}
\end{equation}
\\
Using (\ref{eq48})-(\ref{eq50}), $\dot{V}$ can be expressed as (\ref{eq51}),
\begin{equation}\label{eq51}
    \dot{V}\leq -\rho(t) V^{\frac{1}{2}}
\end{equation}
and it can be concluded that solutions of (\ref{eq29}) converge to the ultimate bound $\epsilon^*$ in finite time as long as the function $\rho(t) = \dfrac{\lambda_{min}(\tilde{Q})\sqrt{\lambda_{min}(P)}}{\lambda_{max}(P)}>0$.

\theoremstyle{remark}
\begin{remark} When the
Lyapunov function is continuously differentiable, or at least locally Lipschitz continuous, differentiating it along the trajectories is straightforward. However,  here the Lyapunov function used for the convergence analysis fails that condition due the term $s$/$\norm{s}^\frac{1}{2}$. In this case, the negative definiteness of $\dot{V}$ is satisfied almost everywhere except at $s=0$. But the energy function $V$ monotonically decreases along the trajectories and hence the point of non-differentiability doesn't create any problem and can be proved using the theorem of Zubov \cite[Theorem ~20.2, p.~568]{poz}.
\end{remark}
\section{SIMULATION STUDIES}\label{Simulation Studies}
Here, the performance of the proposed controller is compared with the existing time delayed impedance controller \cite{jung2021similarity} to study its effectiveness. Cartesian space dynamics of a two link manipulator was derived from the joint space dynamics using relations (\ref{eq5})-(\ref{eq7}) with link masses, $m_1=0.8kg, m_2=0.7kg$, link lengths $l_1=0.6m, l_2=0.5m$, and $g=9.8{m}{s^{-2}}$. Subsequent simulations were carried out in the Cartesian space, where the reference Cartesian position trajectories were selected as ${x_d}(t) = 0.1-0.1\cos(2t)$, ${y_d}(t) = 0.35-0.1\cos(2t)$ with an environment location, ${x_e}(t) = 0.05-0.1\cos(2t)$ and ${y_e}(t) = 0.3-0.1\cos(2t)$.
\begin{equation}\label{eq52}
     {M(q)} = \begin{bmatrix}
{M}_{11} &  {M}_{12}\\[.25cm]
{M}_{21} &  {M}_{22}
\end{bmatrix}
\end{equation}

$M_{11}=(m_1+m_2)l_1^2+m_2l_2(l_2+2l_1\cos(q_2))$,\\ \\
$M_{12}=M_{21}=m_2l_2(l_2+l_1\cos(q_2))$, $M_{22} = m_2l_2^2$
\begin{equation}\label{eq53}
\begin{split}
{C(q,\dot{q})} = \begin{bmatrix}
-m_2l_1l_2\sin(q_2)\dot{q}_2 & -m_2l_1l_2\sin(\dot{q}_2+\dot{q}_1)\\[.25cm]
0 & m_2l_1l_2\sin(q_2)\dot{q}_2\end{bmatrix}
\end{split}
\end{equation}
\begin{equation}\label{eq54}
\begin{split}
    {g(q)} = \begin{bmatrix}m_1l_1g\cos(q_1)+m_2g(l_2\cos(q_1+q_2)\\+l_1\cos(q_1))\vspace{0.2cm}\\m_2gl_2\cos(q_1+q_2)\end{bmatrix}
\end{split}
\end{equation}
\begin{equation}\label{eq55}
    {d_x}(t) = \begin{bmatrix}0.5\sin(t)\\0.5\sin(t)\end{bmatrix}
\end{equation}
\begin{table}[h!]
\caption{Control Parameters For Proposed and Existing Method}
\label{Control Parameters}
\begin{center}
\begin{tabular}{c c c c}
\hline
\hline
Parameter & Numerical Value & TDC Parameters & Numerical Value\\[0.1cm]
\hline\\
$\Gamma_1$ & 10$I_{2}$ & $\bar{M}_x$ & 0.01$I_{2}$\\[0.1cm]
$\Gamma_2$ & 10$I_{2}$ & $h$ & 5ms\\[0.1cm]
$K_m$ & 60$I_{2}$ & $\gamma_{01}$ & 15\\[0.1cm]
$D_m$ & 35$I_{2}$ & $\gamma_{02}$ & 0.1 \\[0.1cm]
$M_m$ & $I_{2}$ & $\gamma_{11}$ & 0.03\\[0.1cm]
$K_e$  & 50$I_2$ & $\gamma_{12}$ & 0.03\\[0.1cm]
\hline
\hline
\multicolumn{3}{l}{$^{1}$\footnotesize{$I_{2}$ denotes a second order identity matrix}}
\end{tabular}
\end{center}
\end{table}
\subsection{Discussion on Results}
\subsubsection{Impedance Tracking ($F_e\neq0$)}
\begin{figure*}[h!]
\centering 
      \subfigure[Position error in y-axis\label{fig2:a} ($m$)]{\includegraphics[width=.329\textwidth]{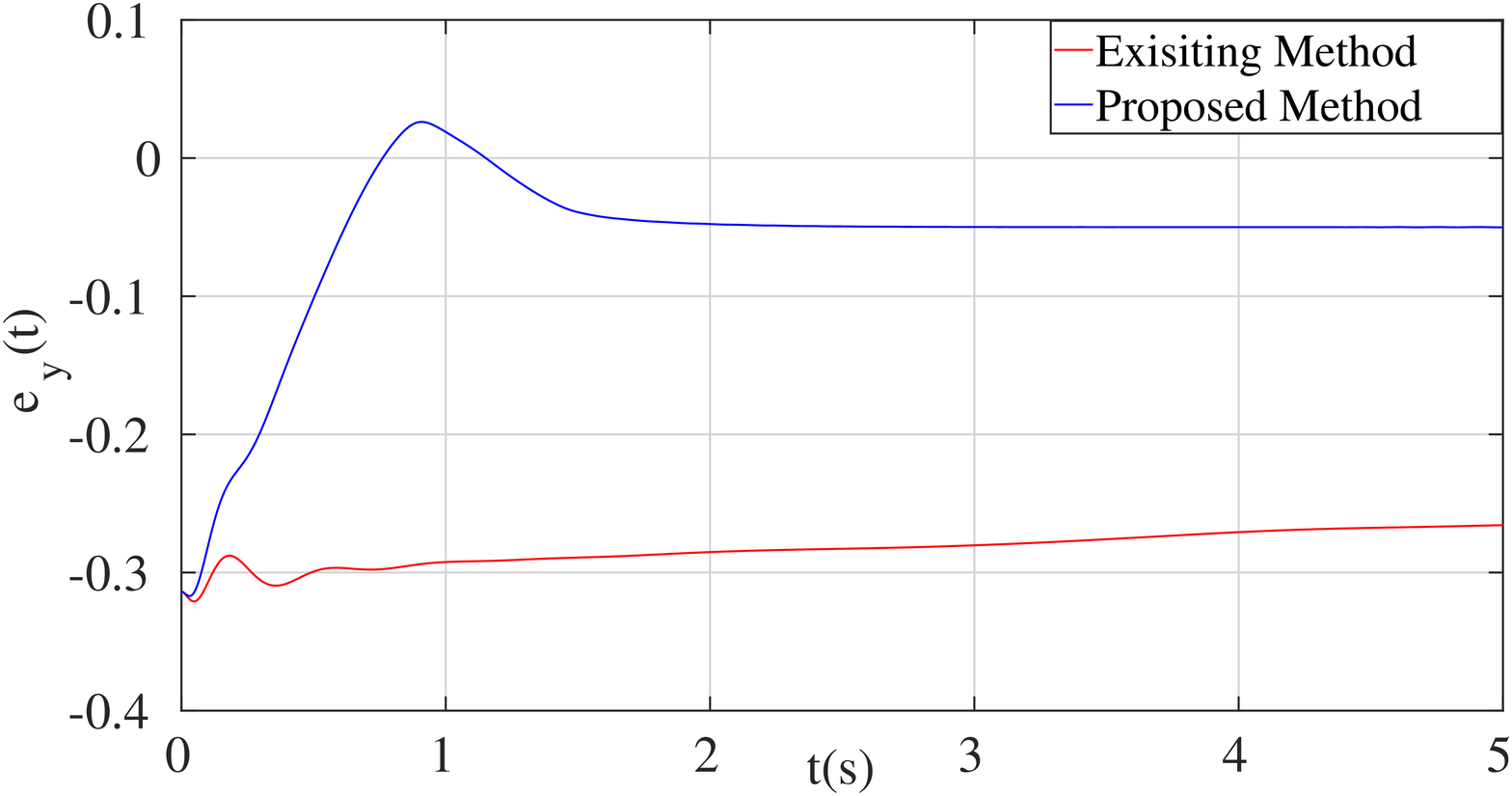}}
      \subfigure[Trajectory tracking \label{fig2:b}        (m)]{\includegraphics[width=.329\textwidth]{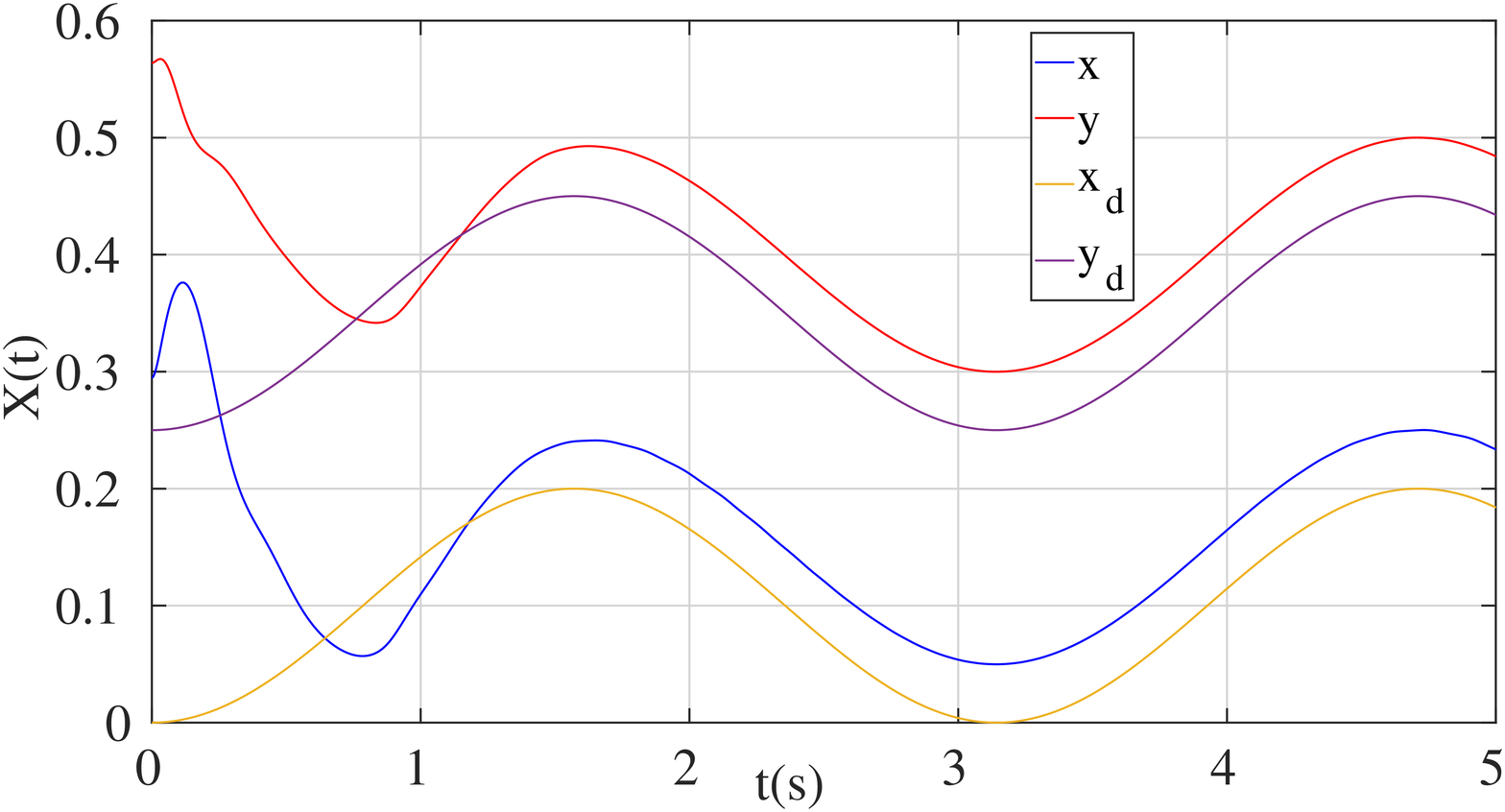}}
      \subfigure[End effector forces (N)\label{fig2:c} ]{\includegraphics[width=.329\textwidth]{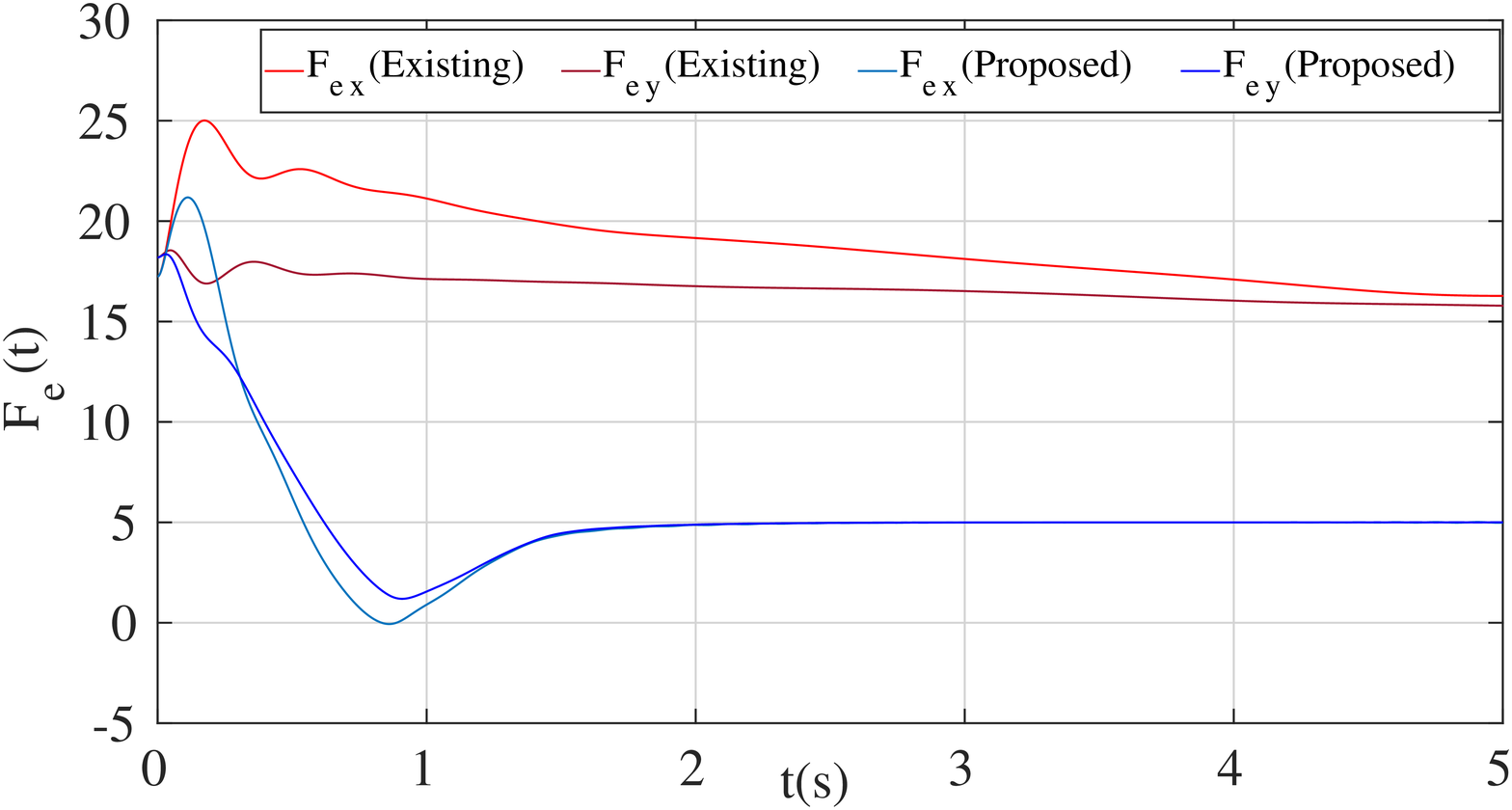}}
      \caption{Results with Proposed Robust Artificial Delay based Impedance Controller for Interaction Force $F_e\neq0$}
       \label{fig2}
      \end{figure*}

When the end effector is subjected to interaction forces, $\Delta a$ tries to ensure that both $s$, $\dot{s}$ converge to a bound in finite time by negotiating the TDE error $\sigma$. Convergence of the variable $\dot{s}$, i.e., $\eta$ to the desired bound implies that the convergence to the desired impedance model is achieved. It can be observed from fig.\ref{fig1:a} and fig.\ref{fig1:b} that the components of the impedance model error $\eta_x$ and $\eta_y$ reach steady state in finite time with the proposed controller whereas the existing time delayed impedance controller \cite{jung2021similarity} is not able to achieve the same. Also, from fig.\ref{fig1:c}, fig.\ref{fig2:a} and fig.\ref{fig2:b} it is clear that the errors $e_x, e_y$ attain a non zero value in steady state as $\eta$ converges to its corresponding steady state which is expected in impedance control where an equilibrium position between the environment location and the reference is achieved leading to an indirect application of force on the compliant environment as seen from fig.\ref{fig2:c}.
      \begin{figure*}
      \centering
      \subfigure[Trajectory tracking\label{fig3:a} (m)]{\includegraphics[width=.329\textwidth]{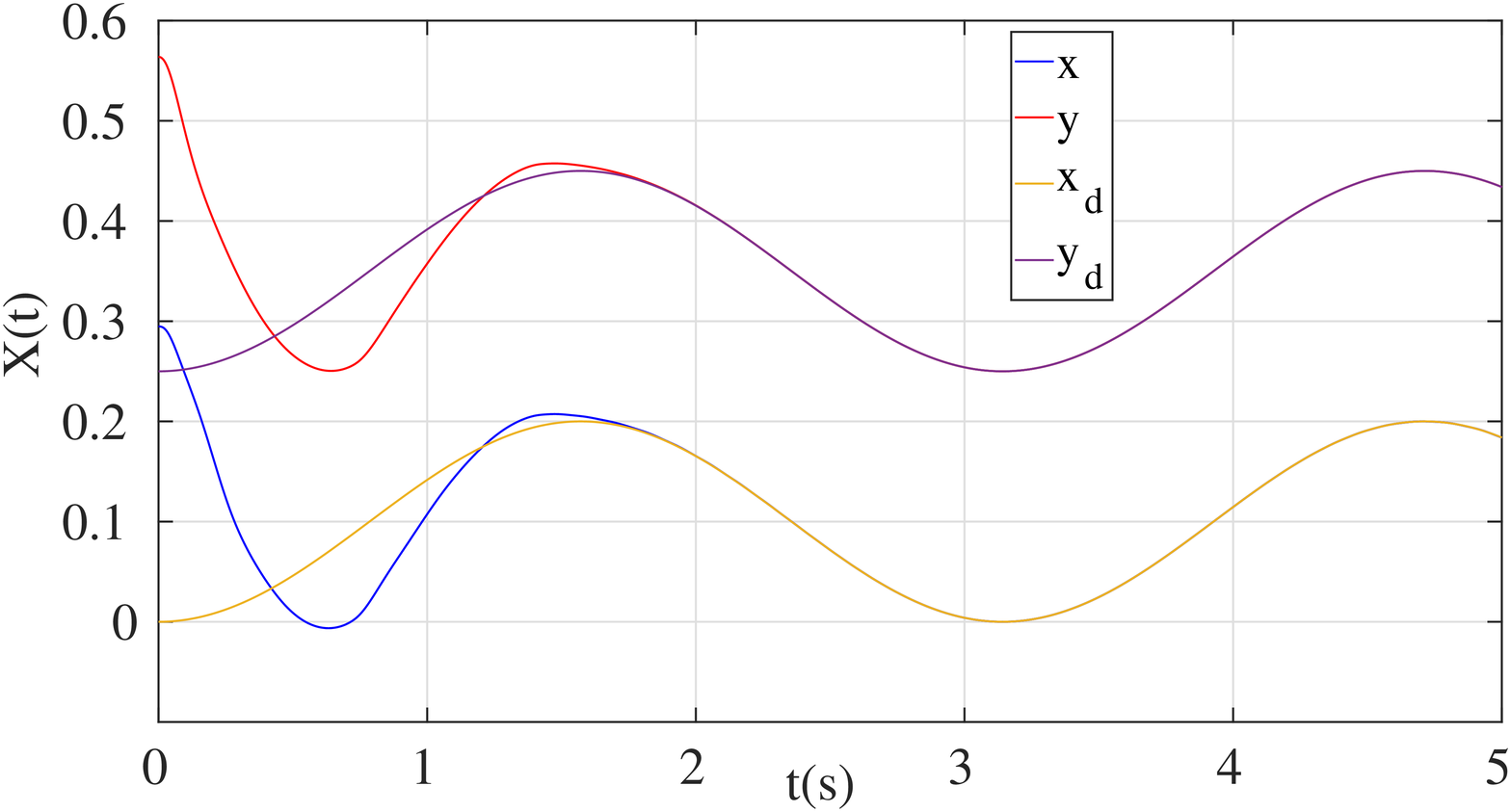}}
      \subfigure[Position error in x-axis and y-axis \label{fig3:b} ($m$)]{\includegraphics[width=.329\textwidth]{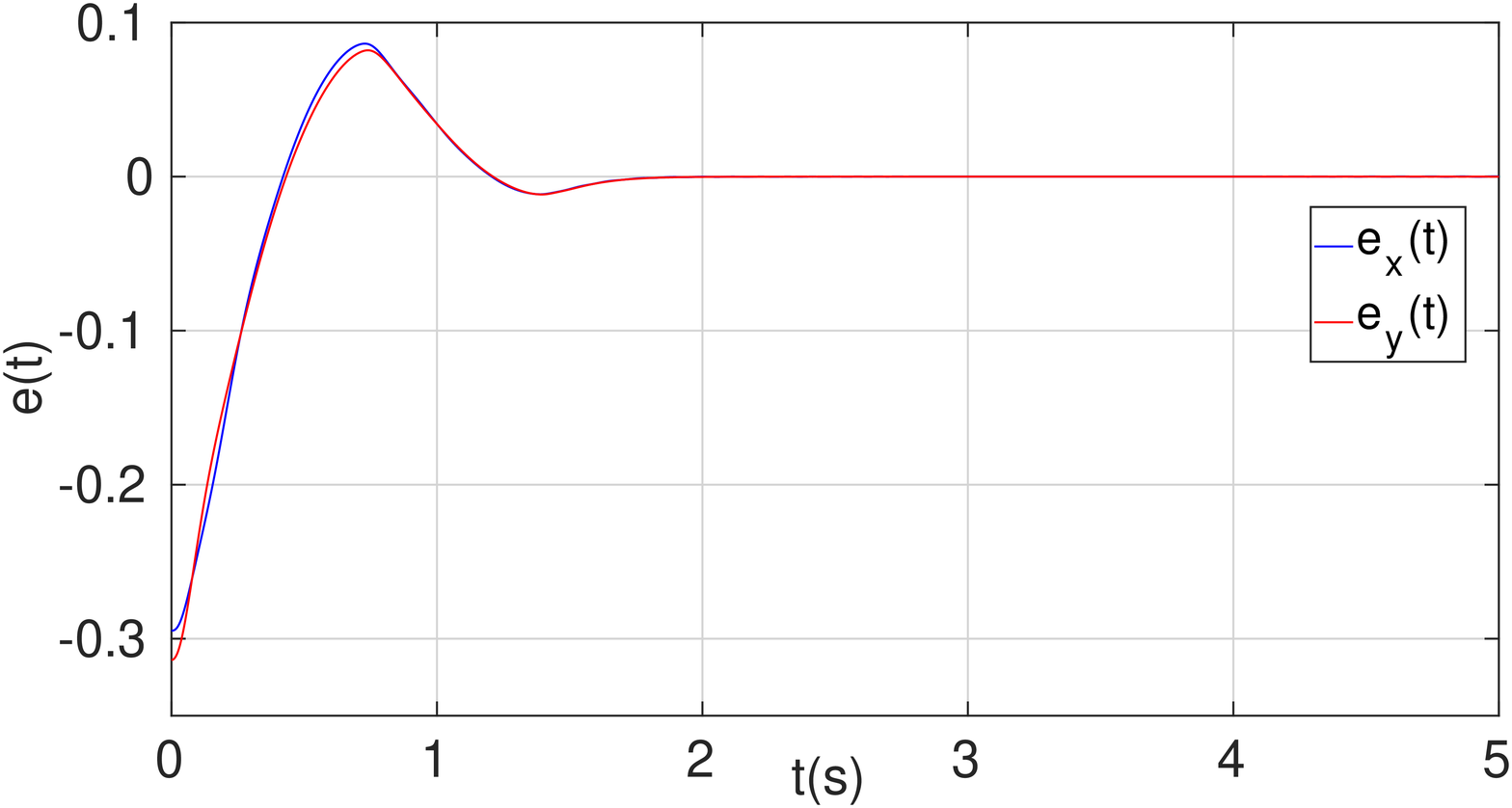}}
          \subfigure[Joint torques \label{fig3:c} (N-m)]{\includegraphics[width=.329\textwidth]{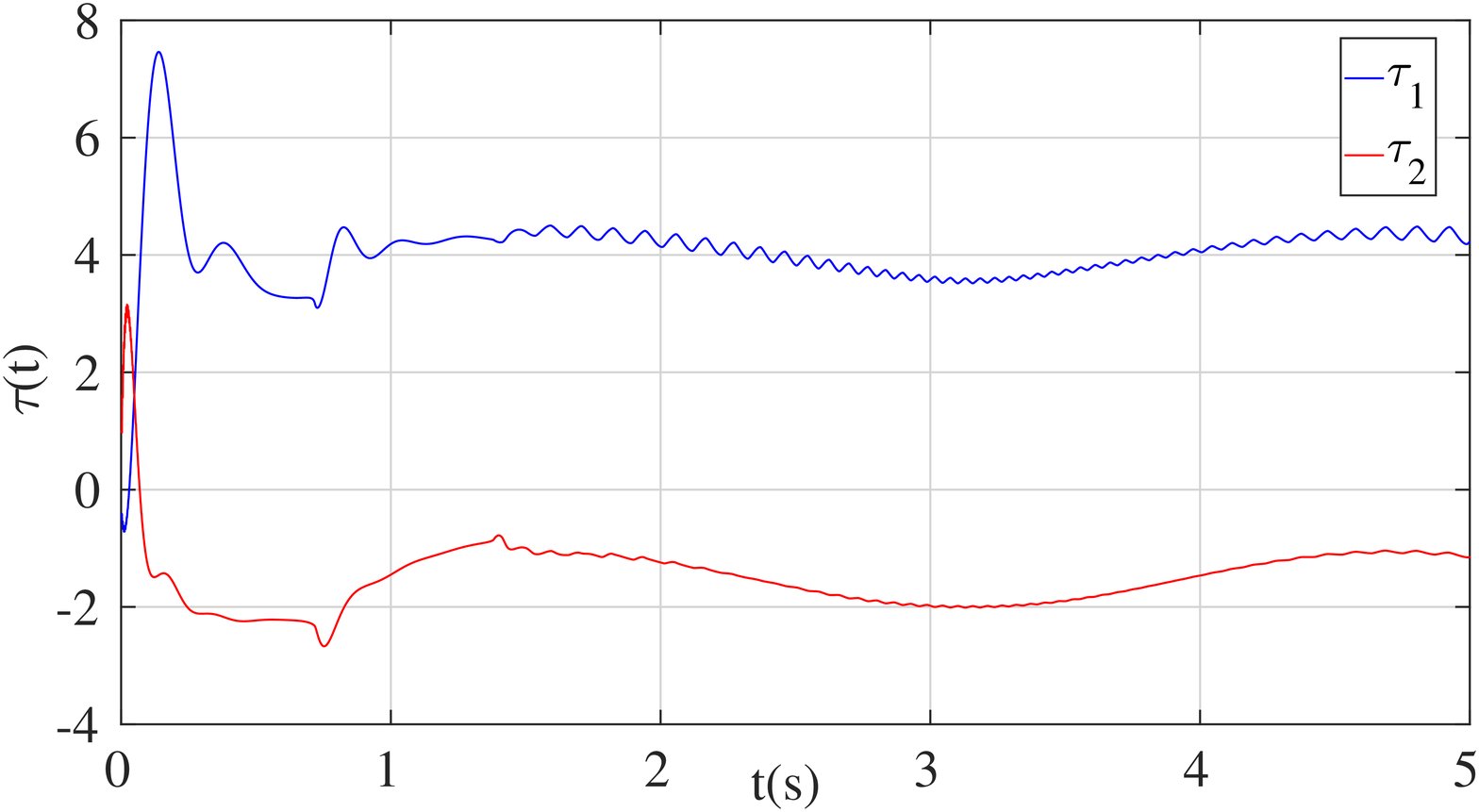}}
       \caption{Results with Proposed Robust Artificial Delay based Impedance Controller for Interaction Force $F_e=0$}
       \label{fig2}
      \end{figure*}

\subsubsection{Position Tracking ($F_e=0$)}
During free motion of the robot, the interaction force vanishes from the dynamics and this simplifies $\eta$ in (\ref{eq14}) to the second order error dynamic model (\ref{eq56}) where the gains $K_p, K_d>0$. Here, $\Delta a$ will ensure convergence of the errors $e_x, e_y$ to a bound by compensating $\sigma$ as shown in fig\ref{fig3:a} and fig\ref{fig3:b}.
\begin{equation}
\label{eq56}
{\eta}={\ddot{e}+K_d\dot{e}+K_pe}
\end{equation}
The continuous structure of $\Delta a$ as seen in fig\ref{fig3:c} provides smooth joint torques for practical implementation.
\section{Conclusions}\label{Conclusions}
 This article proposes an artificial delay based impedance controller for robotic manipulators with an STC inspired outer loop modification for TDE error mitigation. The robot dynamics consisting of interaction forces is first approximated using the past measurement data of input and state and then the estimation error is compensated with the proposed second order switching controller. Moreover, the use of such control law facilitated the design with continuous control effort which was unlikely in case of simple switching control law. The robust modification could successfully handle the uncertainties involved which can be substantiated from the simulation results. Considering more general interaction forces is under further investigation. 





\bibliographystyle{IEEEtran}
\bibliography{bibitem}

\begin{thebibliography}{10}
\providecommand{\url}[1]{#1}
\csname url@samestyle\endcsname
\providecommand{\newblock}{\relax}
\providecommand{\bibinfo}[2]{#2}
\providecommand{\BIBentrySTDinterwordspacing}{\spaceskip=0pt\relax}
\providecommand{\BIBentryALTinterwordstretchfactor}{4}
\providecommand{\BIBentryALTinterwordspacing}{\spaceskip=\fontdimen2\font plus
\BIBentryALTinterwordstretchfactor\fontdimen3\font minus
  \fontdimen4\font\relax}
\providecommand{\BIBforeignlanguage}[2]{{%
\expandafter\ifx\csname l@#1\endcsname\relax
\typeout{** WARNING: IEEEtran.bst: No hyphenation pattern has been}%
\typeout{** loaded for the language `#1'. Using the pattern for}%
\typeout{** the default language instead.}%
\else
\language=\csname l@#1\endcsname
\fi
#2}}
\providecommand{\BIBdecl}{\relax}
\BIBdecl

\bibitem{gasparetto2012automatic}
A.~Gasparetto, R.~Vidoni, D.~Pillan, and E.~Saccavini, ``Automatic path and
  trajectory planning for robotic spray painting,'' in \emph{ROBOTIK 2012; 7th
  German Conference on Robotics}.\hskip 1em plus 0.5em minus 0.4em\relax VDE,
  2012, pp. 1--6.

\bibitem{raibert1981hybrid}
M.~H. Raibert and J.~J. Craig, ``Hybrid position/force control of
  manipulators,'' 1981.

\bibitem{mason1981compliance}
M.~T. Mason, ``Compliance and force control for computer controlled
  manipulators,'' \emph{IEEE Transactions on Systems, Man, and Cybernetics},
  vol.~11, no.~6, pp. 418--432, 1981.

\bibitem{stepien1987control}
T.~Stepien, L.~Sweet, M.~Good, and M.~Tomizuka, ``Control of tool/workpiece
  contact force with application to robotic deburring,'' \emph{IEEE Journal on
  Robotics and Automation}, vol.~3, no.~1, pp. 7--18, 1987.

\bibitem{whitney1977force}
D.~E. Whitney, ``Force feedback control of manipulator fine motions,'' 1977.

\bibitem{colgate1989analysis}
E.~Colgate and N.~Hogan, ``An analysis of contact instability in terms of
  passive physical equivalents,'' in \emph{Proceedings, 1989 international
  conference on robotics and automation}.\hskip 1em plus 0.5em minus
  0.4em\relax IEEE, 1989, pp. 404--409.

\bibitem{hogan1984impedance}
N.~Hogan, ``Impedance control: An approach to manipulation,'' in \emph{1984
  American control conference}.\hskip 1em plus 0.5em minus 0.4em\relax IEEE,
  1984, pp. 304--313.

\bibitem{hogan1985impedance}
------, ``Impedance control: An approach to manipulation: Part
  ii—implementation,'' 1985.

\bibitem{chan1991robust}
S.~Chan, B.~Yao, W.~Gao, and M.~Cheng, ``Robust impedance control of robot
  manipulators.'' \emph{International Journal of Robotics \& Automation},
  vol.~6, no.~4, pp. 220--227, 1991.

\bibitem{ibeas2004robust}
A.~Ibeas and M.~De~la Sen, ``Robust impedance control of robotic
  manipulators,'' in \emph{2004 43rd IEEE Conference on Decision and Control
  (CDC)(IEEE Cat. No. 04CH37601)}, vol.~2.\hskip 1em plus 0.5em minus
  0.4em\relax IEEE, 2004, pp. 1258--1263.

\bibitem{kelly1989adaptive}
R.~Kelly, R.~Carelli, M.~Amestegui, and R.~Ortega, ``On adaptive impedance
  control of robot manipulators,'' in \emph{Proceedings, 1989 International
  Conference on Robotics and Automation}.\hskip 1em plus 0.5em minus
  0.4em\relax IEEE, 1989, pp. 572--577.

\bibitem{lu1991impedance}
W.-S. Lu and Q.-H. Meng, ``Impedance control with adaptation for robotic
  manipulations,'' \emph{IEEE Transactions on Robotics and Automation}, vol.~7,
  no.~3, pp. 408--415, 1991.

\bibitem{zhang2016development}
T.~Zhang, L.~Jiang, S.~Fan, X.~Wu, and W.~Feng, ``Development and experimental
  evaluation of multi-fingered robot hand with adaptive impedance control for
  unknown environment grasping,'' \emph{Robotica}, vol.~34, no.~5, pp.
  1168--1185, 2016.

\bibitem{colbaugh1993direct}
R.~Colbaugh, H.~Seraji, and K.~Glass, ``Direct adaptive impedance control of
  robot manipulators,'' \emph{Journal of Robotic Systems}, vol.~10, no.~2, pp.
  217--248, 1993.

\bibitem{roy2020adaptive}
S.~Roy and I.~N. Kar, \emph{Adaptive-Robust Control with Limited Knowledge on
  Systems Dynamics}.\hskip 1em plus 0.5em minus 0.4em\relax Springer, 2020.

\bibitem{souzanchi2017robust}
M.~Souzanchi-K, A.~Arab, M.-R. Akbarzadeh-T, and M.~M. Fateh, ``Robust
  impedance control of uncertain mobile manipulators using time-delay
  compensation,'' \emph{IEEE Transactions on Control Systems Technology},
  vol.~26, no.~6, pp. 1942--1953, 2017.

\bibitem{jung2021similarity}
S.~Jung and J.~W. Lee, ``Similarity analysis between a nonmodel-based
  disturbance observer and a time-delayed controller for robot manipulators in
  cartesian space,'' \emph{IEEE Access}, vol.~9, pp. 122\,299--122\,307, 2021.

\bibitem{roy2019new}
S.~Roy, J.~Lee, and S.~Baldi, ``A new continuous-time stability perspective of
  time-delay control: Introducing a state-dependent upper bound structure,''
  \emph{IEEE Control Systems Letters}, vol.~3, no.~2, pp. 475--480, 2019.

\bibitem{roy2017adaptive}
S.~Roy, I.~N. Kar, J.~Lee, and M.~Jin, ``Adaptive-robust time-delay control for
  a class of uncertain euler--lagrange systems,'' \emph{IEEE Transactions on
  Industrial Electronics}, vol.~64, no.~9, pp. 7109--7119, 2017.

\bibitem{kali2018super}
Y.~Kali, M.~Saad, K.~Benjelloun, and C.~Khairallah, ``Super-twisting algorithm
  with time delay estimation for uncertain robot manipulators,''
  \emph{Nonlinear Dynamics}, vol.~93, no.~2, pp. 557--569, 2018.

\bibitem{tran2021novel}
M.-T. Tran, D.-H. Lee, S.~Chakir, and Y.-B. Kim, ``A novel adaptive
  super-twisting sliding mode control scheme with time-delay estimation for a
  single ducted-fan unmanned aerial vehicle,'' in \emph{Actuators}, vol.~10,
  no.~3.\hskip 1em plus 0.5em minus 0.4em\relax Multidisciplinary Digital
  Publishing Institute, 2021, p.~54.

\bibitem{shtessel2014sliding}
Y.~Shtessel, C.~Edwards, L.~Fridman, A.~Levant \emph{et~al.}, \emph{Sliding
  mode control and observation}.\hskip 1em plus 0.5em minus 0.4em\relax
  Springer, 2014, vol.~10.

\bibitem{poz}
A.~S. Poznyak, \emph{Advanced Mathematical Tools for Automatic Control
  Engineers: Deterministic Techniques}.\hskip 1em plus 0.5em minus 0.4em\relax
  Oxford: Elsevier, 2008.

\end{thebibliography}

\end{document}